\documentclass[preprint,12pt,authoryear]{elsarticle}

\usepackage{amssymb}
\usepackage{amsmath}
\usepackage{graphicx}
\usepackage{subcaption}
\usepackage{tabularx}
\usepackage{booktabs}
\usepackage{hyperref}
\usepackage{multirow}

\begin{document}

\begin{frontmatter}



\title{A Distributed Hybrid Quantum Convolutional Neural Network for Medical Image Classification} 


\author[1]{Yangyang Li\corref{cor1}}
\cortext[cor1]{Corresponding author}
\ead{yyli@xidian.edu.cn}
\author[1]{Zhengya Qi}
\author[1]{Yuelin Li}
\author[1]{Haorui Yang}
\author[1]{Ronghua Shang}
\author[1]{Licheng Jiao}
\affiliation[1]{organization={Key Laboratory of Intelligent Perception and Image Understanding of Ministry of Education, Joint International Research Laboratory of Intelligent Perception and Computation, International Research Center for Intelligent Perception and omputation, Collaborative Innovation Center of Quantum Information of Shaanxi
Province, School of Artificial Intelligence, Xidian University},
            addressline={No. 2 Taibai South Road, Yanta District},
            city={Xi’an},
            postcode={710071},
            state={Shaanxi},
            country={China}}

\begin{abstract}

Medical images are characterized by intricate and complex features, requiring interpretation by physicians with medical knowledge and experience. 
Classical neural networks can reduce the workload of physicians, but can only handle these complex features to a limited extent. 
Theoretically, quantum computing can explore a broader parameter space with fewer parameters, but it is currently limited by the constraints of quantum hardware.
Considering these factors, we propose a distributed hybrid quantum convolutional neural network based on quantum circuit splitting. This model leverages the advantages of quantum computing to effectively capture the complex features of medical images, enabling efficient classification even in resource-constrained environments. Our model employs a quantum convolutional neural network (QCNN) to extract high-dimensional features from medical images, thereby enhancing the model's expressive capability.
By integrating distributed techniques based on quantum circuit splitting, the 8-qubit QCNN can be reconstructed using only 5 qubits.
Experimental results demonstrate that our model achieves strong performance across 3 datasets for both binary and multiclass classification tasks. 
Furthermore, compared to recent technologies, our model achieves superior performance with fewer parameters, and experimental results validate the effectiveness of our model.

\end{abstract}

\begin{keyword}
Medical image classification\sep Quantum convolutional neural network\sep Quantum circuit splitting\sep Distributed quantum computing

\end{keyword}
\end{frontmatter}

\section{Introduction}
\label{sec1}

Medical images play a crucial role in clinical diagnosis, characterized by intricate and complex features. Accurate interpretation of these images is imperative, as diagnostic outcomes directly influence patient treatment plans. However, this task demands a high level of medical expertise and extensive experience, imposing significant demands and workload on healthcare professionals.

To facilitate more efficient and accurate disease diagnosis while alleviating the workload of physicians, deep learning techniques have been incorporated into the field of medical image classification (\cite{albahri2020systematic}). 
Neural networks are effectively used to process these complex data, assisting in the extraction of valuable information and features, accelerating the processing and analysis of medical images, enhancing efficiency, and ultimately improving diagnostic accuracy and speed.
However, models such as convolutional neural networks (CNNs)(\cite{jiwani2022convolutional}), when processing large volumes of medical images, must increase model parameters and deepen network layers to extract more features, which inevitably increase the computational complexity of the network.

With the rapid advancement of quantum computing, quantum computers have demonstrated advantages in tackling challenging problems that are difficult for classical computers to solve (\cite{de2021materials}). 
Leveraging quantum parallelism to swiftly process large volumes of medical data, the properties of quantum superposition and entanglement (\cite{schuld2017implementing}) enable more efficient high-dimensional data extraction using fewer quantum computing resources.
To fully harness the acceleration advantages of quantum computing for real-world issues, certain requirements must be met regarding the quantity and quality of qubits. 

Unfortunately, many of the quantum computing resources required for these technologies exceed the capabilities of today's quantum computers.
The limitations of quantum computing hardware, particularly regarding entanglement operations and the number of logical qubits (\cite{klimov2018fluctuations}), have rendered distributed quantum computing an important strategy for addressing the problem of qubit expansion and achieving quantum advantage. 
However, research on distributed quantum neural networks is noticeably lacking and requires further development.

Based on the aforementioned background,  a distributed hybrid quantum convolutional neural network is proposed. 
This approach aims to leverage the strengths of both quantum and classical computing to improve the efficiency and accuracy of medical image classification.
By introducing specially designed a quantum convolutional neural network, it effectively extracts image features in multidimensional space. 
The distribution of the quantum neural network is achieved through quantum circuit splitting, which allows the required 8-qubit circuit to be reconstructed using a set of 4-qubit and 5-qubit circuits.
This allows the execution of an 8-qubit QCNN using only a 5-qubit quantum computer, significantly reducing the spatial complexity of quantum computing and conserving quantum resources.

The main contributions of this paper ares as follows:
\begin{itemize}

\item For the first time, building on quantum circuit splitting, a novel distributed hybrid quantum convolutional neural network is proposed for the classification of medical images.

\item A quantum convolutional neural network is introduced to enhance the model's understanding of both fine-grained and global features in medical images. High accuracy is achieved, and excellent performance is demonstrated.

\item By employing quantum circuit splitting techniques, an 8-qubit QCNN can be expressed using only 5 qubits, nearly halving the required number of qubits. This approach significantly reduces the complexity of implementing quantum circuits while maintaining excellent performance.

\item Compared to several recent models, our work achieves superior performance while utilizing fewer parameters.

\end{itemize}

The structure of this paper is organized as follows: related work is reviewed in \autoref{Related works}. A detailed description of the model architecture is provided in \autoref{Method}. An evaluation of the performance of the model and comparisons with other networks are presented in \autoref{Experiments}. Finally, the paper is concluded and future directions are discussed in \autoref{Conclusions and future work}.

\section{Related Works}
\label{Related works}

\subsection{CNN for medical image Classification}

Convolutional neural networks, as a significant branch of deep learning, are utilized in medical image processing.
CNN models such as GoogleNet (\cite{7298594}), AlexNet (\cite{krizhevsky2017imagenet}), and Inception (\cite{dong2020inception}) have been demonstrated to perform outstandingly in image classification and recognition tasks (\cite{li2024attention}). In 2018, features were extracted using a pre-trained AlexNet and combined with an ECOC SVM classifier for skin cancer classification. In 2021, an automated network for brain tumor classification using SVM and CNN was proposed by \cite{deepak2021automated}. In 2023, a method that combines GoogleNet with the Dynamic Dipper Throated Optimization Algorithm (DDTPSO) for breast cancer classification was employed by \cite{alhussan2023breast}

\subsection{Quantum convolutional neural network}

Recent successes in classical neural networks, coupled with advancements in quantum computing, have inspired the investigation of interactions between these technologies, culminating in the development of quantum neural networks.

The idea of quantum neural computation was first introduced by \cite{kak1995quantum}.
In 2019, a physically realizable quantum convolutional neural network was introduced by \cite{cong2019quantum}, resembling the structure of convolutional neural networks, with network layers constructed from parameterized quantum circuits.
QCNNs were employed by \cite{qu2023qnmf} to extract features from medical images and integrate them with other modal features. 
In 2024, a custom convolutional neural network deep learning model was proposed by \cite{rao2024hybrid}, built on a quantum variational circuit with encoding, entanglement, and measurement capabilities.

Hybrid models have been proposed by researchers (\cite{liu2021hybrid}), building on the strengths of both classical neural networks and quantum neural networks.
In the work by \cite{liang2021hybrid}, a hybrid quantum-classical convolutional neural network is presented, which integrates residual block structures with quantum convolutional neural networks.
In 2023, a hybrid quantum convolutional neural network was constructed by (\cite{qu2023iomt}) to extract temporal features from electrocardiogram signals for detecting abnormal heartbeats. 

Despite the rapid development of QCNNs, their application in medical image analysis remains limited. Therefore, designing a QCNN tailored for medical image classification is crucial.

\subsection{Distributed quantum computing}

Quantum computing for medical image classification is typically limited by qubit constraints due to hardware limitations. A promising solution to this challenge is distributed quantum computing (DQC), which offers greater feasibility and flexibility.

In 1993, DQC based on quantum teleportation was proposed by \cite{bennett1993teleporting}. 
In 2000, fault-tolerant quantum logic gates, were constructed by \cite{zhou2000methodology} from Stanford University using a method similar to quantum teleportation.

Among these, quantum circuit splitting is favored due to its ability to more accurately reconstruct quantum states and its broader applicability.
In 2020, quantum circuit splitting was proposed and demonstrated by \cite{peng2020simulating}.
In 2022, quantum circuit splitting was utilized by \cite{eddins2022doubling} to express a 2\textit{n}-qubit quantum simulation as a linear combination of two \textit{n}-qubit quantum simulations, effectively doubling the problem size that the chip can handle. 
In 2024, as described by \cite{harrow2024optimal}, by employing quantum circuit splitting, involving the measurement of the existing quantum state and re-preparation based on the measurement results, facilitating the decomposition of the quantum circuit.

\section{Method}
\label{Method}  

\begin{figure}[htbp]
    \centering
    \includegraphics[width=1\linewidth]{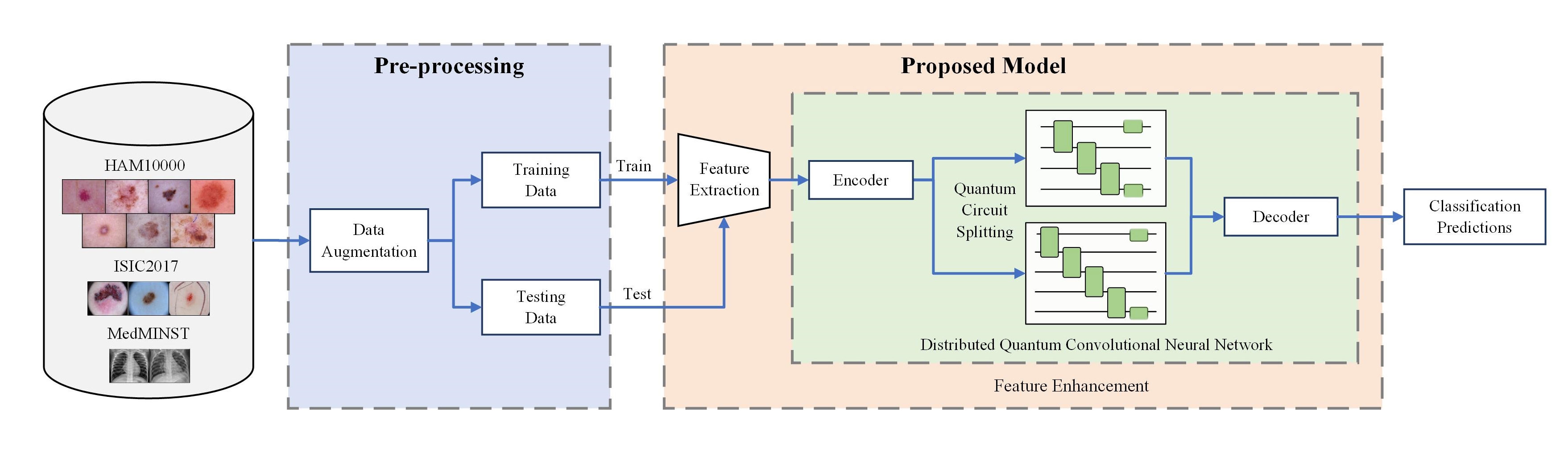}
    \caption{Complete workflow of medical image classification model}\label{workflow}
\end{figure}

\begin{figure}[htbp]
    \centering
    \includegraphics[width=1\linewidth]{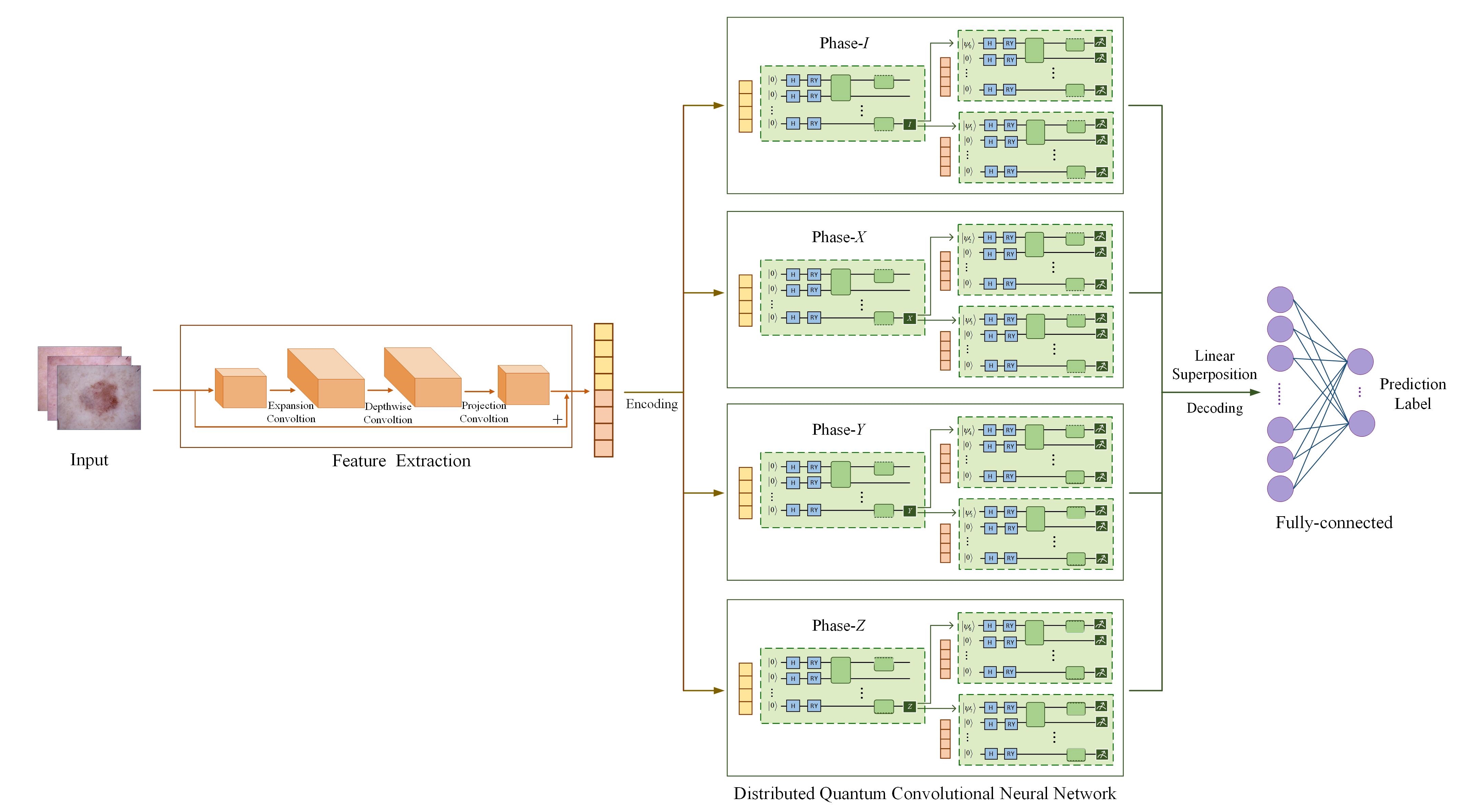}
    \caption{The framework of the proposed model}\label{framework}
\end{figure}

The execution flow of the system is illustrated in \autoref{workflow}.
In the pre-processing phase, data augmentation and the division of the dataset into training and testing sets are performed. 
The trained model is obtained using the training data, and the model's performance is evaluated on the testing data based on classification predictions. 
The proposed model first employs a CNN for feature extraction and dimensionality reduction. 
The extracted feature vector is then used as input for a distributed 8-qubit quantum convolutional neural network. 
The 8-qubit quantum convolutional neural network employs the quantum circuit splitting technique, consisting of four groups, each formed by one 5-qubit QCNN and one 4-qubit QCNN.
Finally, a fully connected layer is utilized for classification tasks.
The framework of the proposed model is depicted in \autoref{framework}.

\subsection{Feature extraction}

A lightweight CNN architecture, MobileNetV2 \cite{sandler2018mobilenetv2}, was employed to leverage its ability to capture spatial hierarchical features in images for medical image feature extraction.
This network introduces "Inverted Residuals and Linear Bottlenecks," which implement an inverse residual structure opposite to that of traditional residual networks. 
Specifically, dimensions are first expanded through a \(1\times1\) convolution, followed by feature extraction via depthwise separable convolutions, and concluded with another \(1\times1\) convolution for dimensionality reduction. 
Additionally, the concept of linear bottlenecks is incorporated into MobileNetV2, using linear transformations instead of nonlinear activation functions within the residual blocks. 
These bottleneck layers have fewer channels at both the input and output stages, while the intermediate expansion layers utilize lightweight depthwise convolutions to filter features. 
This approach significantly reduces computational demands and memory usage, further enhancing the model's efficiency.

\subsection{Quantum convolutional neural network}

The learning task is enhanced by using a quantum convolutional neural network to extract high-dimensional features.
The quantum convolutional neural network can be described using a quantum circuit.
These circuits are composed of three main components: quantum state encoding, quantum entanglement layers, and quantum measurement. 
Quantum state encoding is responsible for loading data into the quantum system, with common encoding methods including basis state encoding, amplitude encoding, and angle encoding (\cite{schuld2018supervised}). 
The quantum entanglement layer is employed for precise manipulation of quantum states, such as adjusting the rotation angle of quantum states via rotation gates.
By arranging multiple quantum gates in a predetermined sequence, the initial quantum state can be transformed into the desired outcome. 
Finally, quantum measurement is used to extract classical information from the quantum state while effectively filtering out irrelevant information.

In our research, angle encoding is utilized to transform data from classical states into quantum states. Let $D\subset {{R}^{n}}$ be a finite set, with each data point represented as $x=({{x}_{1}},\cdot \cdot \cdot ,{{x}_{n}})\in D$. An injective function $f$ is employed to map all subsets of $D$ into a particular Hilbert space ${{H}^{m}}$ , ensuring that for $D'\subset D,f(D')\in {{H}^{m}}$ . For angle encoding, the relationship for any input $D'=({{x}_{0}},{{x}_{1}},\cdot \cdot \cdot ,{{x}_{n-1}})$ is defined as follows:

\begin{equation}\label{eqn-1} 
  f(D')=\frac{1}{C}\sum\limits_{i=1}^{n}{{{x}_{i}}\left| i \right\rangle }
\end{equation}

The basis states of the qubits are represented by $ \left| i \right\rangle$, indicating the different basis vectors of the quantum state. The normalization constant is given by \textit{C}.

\begin{equation}
 C=\sqrt{\sum\limits_{i=1}^{n}{x_{i}^{2}}} 
\end{equation}

In quantum circuits, transformations are commonly achieved using the Hadamard gate (\textit{H} gate) and the \textit{RY} gate. The \textit{H} gate is used to convert qubits from classical states to superposition states, while the \textit{RY} gate is employed to rotate each qubit to the desired quantum state angle based on the input data values. The matrix representations are as follows:

\begin{equation}
H = \frac{1}{\sqrt{2}} \left[ \begin{array}{cc}
1 & 1 \\
1 & -1
\end{array} \right]
\end{equation}

\begin{equation}
RY(\theta) = \exp\left( -i \frac{\theta}{2} Y \right) = \left[ \begin{array}{cc}
\cos\left( \frac{\theta}{2} \right) & -\sin\left( \frac{\theta}{2} \right) \\[0.5em]
\sin\left( \frac{\theta}{2} \right) & \cos\left( \frac{\theta}{2} \right)
\end{array} \right]
\end{equation}

In the quantum entanglement layer, a single-qubit gate (\textit{RZ}) as well as two-qubit controlled rotation gates (\textit{CRY }and \textit{CNOT}) are utilized. In these two-qubit gates, the control qubit is denoted by a black dot, while the target qubit is influenced by the associated operation. 
The matrix representations of these gates are presented as follows:

\begin{equation}
{{R}{Z}}(\phi )=\left[ \begin{array}{cc}
  {{e}^{-i\frac{\phi }{2}}}   &  0\\
    0 & {{e}^{i\frac{\phi }{2}}}
\end{array}\right]
\end{equation}

\begin{equation}
C({{R}{Y}}(\theta ))=\left[ \begin{array}{cc}
I & 0  \\
   0 & {{R}{Y}}(\theta )
\end{array}\right]
\end{equation}

\begin{equation}
\text{CNOT}=\left[ \begin{array}{cccc}
   1 & 0 & 0 & 0  \\
   0 & 1 & 0 & 0  \\
   0 & 0 & 0 & 1  \\
   0 & 0 & 1 & 0  \\
\end{array}\right]
\end{equation}

Inspired by CNNs, QCNNs are composed of convolutional layers and pooling layers.
In the quantum convolutional layer, we employ a quasi-local unitary operator (\textit{U}) applied to the input quantum state in a translationally invariant manner. 
This approach allows for processing and feature extraction from the quantum state with limited depth, akin to the convolution operation in classical convolutional neural networks. 
Consequently, the quantum convolutional network is able to effectively capture local features within quantum states while preserving the consistency of the global structure.

\begin{figure}[h]
    \centering
    \includegraphics[width=0.75\linewidth]{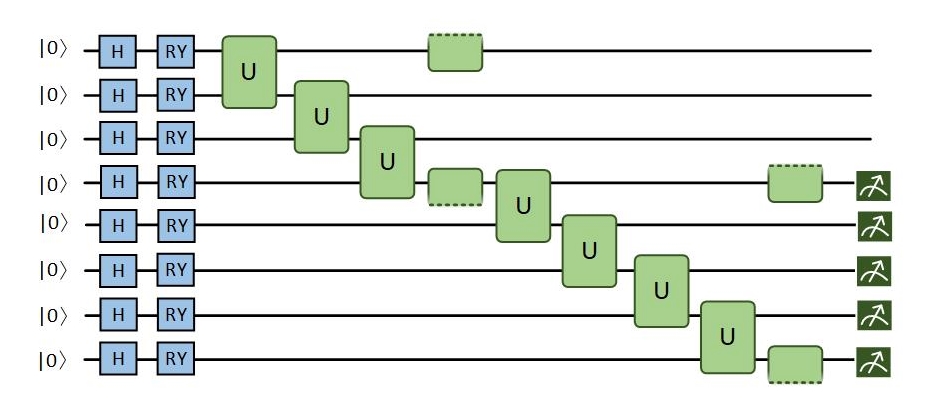}
    \caption{Structure of quantum circuit}\label{circure}
\end{figure}

\begin{figure}[h]
    \centering
    \includegraphics[width=0.5\linewidth]{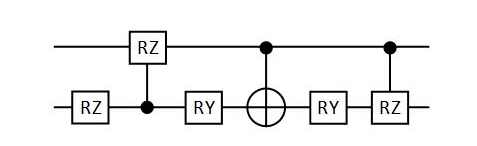}
    \caption{Quantum convolutional kernel}\label{vqe}
\end{figure}

The quantum convolutional layer is designed with a matrix product states (MPS) structure. This structure is characterized by a ladder-like configuration that decomposes high-dimensional tensors into a series of lower-dimensional tensor products.
These lower-rank tensors are interconnected through internal indices, referred to as the bond dimension \textit{D}. The advantage of MPS lies in its ability to approximate complex quantum states with reduced computational complexity. Furthermore, by incorporating strong entanglement layers, the entanglement capabilities of MPS can be enhanced, thereby improving the performance of quantum circuits. \autoref{circure} illustrates the quantum circuit designed in this study.
Each \textit{U} in the MPS circuit is substituted with a simple variational quantum circuit, which represents the convolutional kernel, as illustrated in \autoref{vqe}. 
Subsequently, this circuit can be simulated, implemented on hardware, and optimized in the same manner as other variational quantum circuits. 

The design of quantum pooling layers aims to reduce the size of the feature map.
However, after passing through our quantum convolutional layer, the feature map size has become sufficiently small, allowing for the omission of the pooling layer.

The quantum state is converted into a classical state through measurement, with the results returned as a flat array containing the probabilities of measuring the computational basis states in the current configuration. If the measurement is constrained to \textit{N} measurement circuits, the marginal probabilities can be obtained, resulting in a returned array size of 2\textsuperscript{n}.

\subsection{Quantum circuit splitting}

The current quantum computing hardware is constrained by a limited number of qubits, finite quantum gate operation depths, and restricted connectivity between qubits, making it challenging for large-scale quantum circuits to be handled, particularly when the circuit size exceeds the hardware capabilities. 
By decomposing large circuits into a series of smaller circuits, more complex computational tasks can be performed on existing hardware.

\begin{figure}[htbp]
    \centering
    \begin{subfigure}{0.45\textwidth}
        \centering
        \includegraphics[width=1\linewidth]{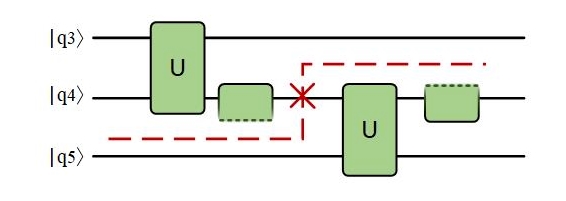}
        \caption{Uncut quantum circuit}
        \label{fig:subfig1}
    \end{subfigure}
    \hfill
    \begin{subfigure}{0.5\textwidth}
        \centering
        \includegraphics[width=1\linewidth]{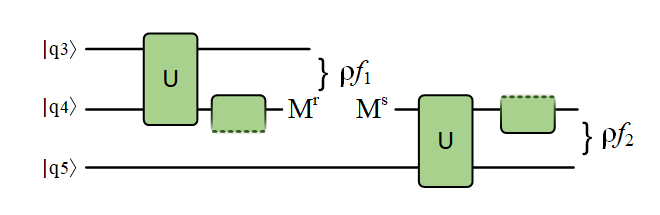}
        \caption{Cut quantum circuit}
        \label{fig:subfig2C}
    \end{subfigure}
    \caption{The quantum circuit diagram before and after cut}
    \label{The quantum circuit diagram before and after cut}
\end{figure}

In the process of quantum circuit splitting, a quantum splitting point is identified, allowing the circuit to be decomposed into two independent segments. To minimize the number of qubits involved, the cut is chosen to be performed in the middle of the quantum circuit. As illustrated in \autoref{The quantum circuit diagram before and after cut}, this approach decomposes an 8-qubit quantum circuit into a linear combination of a 4-qubit circuit and a 5-qubit circuit. 
By performing quantum tomography on the qubits before partitioning, one can learn the quantum state and reconstruct the inferred state in the partitioned qubits. 
Quantum tomography can be viewed as the process of expanding a quantum state and estimating coefficients in a specific basis, here chosen as the Pauli basis. 
By measuring the quantum state for each element in the basis, the quantum state prior to the cut can be reconstructed.
Subsequently, by initializing the downstream qubits associated with the cut to the eigenstates of each element in the basis, the dynamic behavior of the circuit given the initial state can be learned. 
Ultimately, by reweighting the results obtained from multiple initializations, the overall behavior of the uncut circuit can be derived.

The quantum state at the splitting point in this paper can be expressed as follows:
\begin{equation}
\rho =U|{{q}_{3}}{{q}_{4}}{{q}_{5}}\rangle \langle {{q}_{3}}{{q}_{4}}{{q}_{5}}|{{U}^{\dagger }}
\end{equation}

The unitary $\textit{U}$ is considered as arbitrary quantum gates, and ${U}^{\dagger}$ is defined as the Hermitian conjugate of $\textit{U}$. Once the qubit connections between the two unitary modules are severed, the equation can be rewritten as follows:
\begin{equation}
\rho =\frac{1}{2}\sum\limits_{M\in \mathcal{B}}{{{\rho }_{{{f}_{1}}}}}(M)\otimes {{\rho }_{{{f}_{2}}}}(M)\
\label{eq9}
\end{equation}

\begin{equation}
    {{\rho }_{{{f}_{1}}}}(M)=T\text{r}(MU|{{q}_{3}}{{q}_{4}}\rangle \langle {{q}_{3}}{{q}_{4}}|{{U}^{\dagger }})
\end{equation}

\begin{equation}
    {{\rho }_{{{f}_{2}}}}(M)=U(M\otimes |{{q}_{5}}\rangle \langle {{q}_{5}}|){U}^{\dagger}
\end{equation}

where \(M\) is the operator applied to the fourth qubit. Since the Pauli matrices are traceless operators, a spectral decomposition of \(M\) is necessary to obtain a physical interpretation. \autoref{eq9} can then be rewritten as:
\begin{equation}
    \rho =\frac{1}{2}\sum\limits_{M\in \mathcal{B}}{\sum\limits_{r,s=\pm 1}{r}}s{{\rho }_{{{f}_{1}}}}({{M}^{r}})\otimes {{\rho }_{{{f}_{2}}}}({{M}^{s}})
\end{equation}

Here, \(r\) and \(s\) are denoted as the eigenvalues obtained from the spectral decomposition of the operator \(M\), while \({{M}^{r}}\) and \({{M}^{s}}\) represent the corresponding eigenstates. The set \(B\) encompasses the Pauli basis \{\textit{ I, X, Y, Z }\}. The matrix representations of the Pauli operators are as follows: 

\begin{equation}
\begin{array}{cc}
I = \begin{bmatrix} 
1 & 0 \\ 
0 & 1 
\end{bmatrix} & 
X = \begin{bmatrix} 
0 & 1 \\ 
1 & 0 
\end{bmatrix} \\[1.3em]
Y = \begin{bmatrix} 
0 & -i \\ 
i & 0 
\end{bmatrix} & 
Z = \begin{bmatrix} 
1 & 0 \\ 
0 & -1 
\end{bmatrix}
\end{array}
\end{equation}

The information transfer mechanism for the qubit-splitting line involves the measurement of the information stored in the 4th qubit during the computation process. This information is then initialized and transmitted to the remaining part of the circuit.

In a quantum circuit, any quantum operation represented by a unitary matrix \textit{U} can be decomposed into a combination of orthogonal matrices formed by the Pauli basis.
 
\begin{equation}
    U=\frac{Tr(UI)I+Tr(UX)X+Tr(UY)Y+Tr(UZ)Z}{2}
\end{equation}

Based on the following spectral decomposition relationship, the Pauli basis can be rewritten as:

\begin{equation}
 \begin{array}{cc}
I=|0\rangle \langle 0|+|1\rangle \langle 1| & X=|+\rangle \langle +|-|-\rangle \langle -|  \\[0.5em]
Y=|+i\rangle \langle +i|-|-i\rangle \langle -i|  & Z=|0\rangle \langle 0|-|1\rangle \langle 1| 
\end{array}
\end{equation}

Assuming \({{O}_{i}}\) is represented as a Pauli matrix, \({{\psi }_{i}}\) is denoted as the corresponding eigenstates, and \(2{{c}_{i}}\) signifies the associated eigenvalues, the following expression can be derived:

\begin{equation}
     \begin{array}{cc}
{{O}_{1}}=I,\quad {{\psi }_{1}}=|0\rangle \langle 0|,\quad {{c}_{1}}=+\frac{1}{2}; & {{O}_{2}}=I,\quad {{\psi }_{2}}=|1\rangle \langle 1|,\quad {{c}_{2}}=+\frac{1}{2};  \\[0.5em]
   {{O}_{3}}=X,\quad {{\psi }_{3}}=|+\rangle \langle +|,\quad {{c}_{3}}=+\frac{1}{2}; & {{O}_{4}}=X,\quad {{\psi }_{4}}=|-\rangle \langle -|,\quad {{c}_{4}}=-\frac{1}{2};  \\[0.5em]
   {{O}_{5}}=Y,\quad {{\psi }_{5}}=|+\text{i}\rangle \langle +\text{i}|,\quad {{c}_{5}}=+\frac{1}{2}; & 
   {{O}_{6}}=Y,\quad {{\psi }_{6}}=|-\text{i}\rangle \langle -\text{i}|,\quad {{c}_{6}}=-\frac{1}{2};  \\[0.5em]
   {{O}_{7}}=Z,\quad {{\psi }_{7}}=|0\rangle \langle 0|,\quad {{c}_{7}}=+\frac{1}{2}; & {{O}_{8}}=Z,\quad {{\psi }_{8}}=|1\rangle \langle 1|,\quad {{c}_{8}}=-\frac{1}{2}.\\[0.5em]
\end{array}
\end{equation}

From the above expression, it can be observed that an 8-qubit quantum circuit is decomposed into 8 groups, each comprising a 4-qubit circuit and a 5-qubit circuit. As illustrated in \autoref{equ17}, by performing a linear combination of these 8 pairs of dual quantum circuits, the original 8-qubit quantum circuit can be effectively reconstructed.

\begin{equation}
\label{equ17}
    \left\langle {{q}_{4}} \right\rangle =\sum\limits_{m}^{8}{{{c}_{m}}({{O}_{m}}{{\rho }_{m}})}
\end{equation}

\section{Experiments}
\label{Experiments}  

\subsection{Datasets and evaluation metrics}

To train and evaluate the model, three publicly available datasets were utilized: HAM10000 dataset (\cite{tschandl2018ham10000}), ISIC2017 dataset (\cite{8363547}) and MedMNIST (\cite{yang2023medmnist}). 
The first and second datasets are large-scale dermoscopic image multiclass datasets released by the International Skin Imaging Collaboration (ISIC) for the ISIC 2018 and 2017 competitions.
The third dataset is the pneumonia dataset from MedMNIST, which consists of chest X-ray images for binary classification.
The original information of the dataset is shown in \autoref{tab:datasets}.

\begin{table}[h]
\caption{Datasets used in the experiments}
\label{tab:datasets}
    \centering
    \begin{tabular}{ccp{4cm}cc}
        \hline
        Number & Dataset& Categories/Count & Total & Size \\
        \hline
        \multirow{7}{*}{1} & \multirow{7}{*}{HAM10000} & \centering AK: 327 & \multirow{7}{*}{10015} & \multirow{7}{*}{450 $\times$ 600} \\
        & & \centering BCC: 514 & & \\
        & & \centering BKL: 1099 & & \\
        & & \centering DF: 115 & & \\
        & & \centering MEL: 1113 & & \\
        & & \centering NV: 6705 & & \\
        & & \centering VASC: 142 & & \\
        \hline
        \multirow{3}{*}{2} & \multirow{3}{*}{ISIC2017} & \centering BKL: 254 & \multirow{3}{*}{2000} & \multirow{3}{*}{767 $\times$ 1022} \\
        & & \centering MEL: 374 & & \\
        & & \centering NV: 1372 & & \\
        \hline
        \multirow{2}{*}{3} & \multirow{2}{*}{MedMNIST} & \centering NL: 1583 & \multirow{2}{*}{5856} & \multirow{2}{*}{224 $\times$ 224} \\
        & & \centering PNA: 4273 & & \\
        \hline
    \end{tabular}
\end{table}

To address the imbalance in dermoscopic image, data augmentation was applied to the first and second datasets, resulting in 1,372 and 2,000 images for each class, respectively. 
Additionally, data of varying sizes were normalized and cropped to a uniform size of \(224 \times 224\) pixels to ensure consistency and comparability. 

Multiple evaluation metrics were employed to quantify the performance of the experimental results, including AUC, accuracy, recall, precision, f1-score, and specificity (\cite{wu2022skin}). From a clinical perspective, recall and specificity are important because they directly influence accurate identification and exclusion, thereby impacting precise diagnosis and effective patient management.

\subsection{Experimental setup}

In our study, the computational resources were comprised of four NVIDIA GeForce RTX 4090 GPUs, a 192-core Intel CPU, and 128 GB of memory. The model was implemented on the CUDA 12.0 platform using the PyTorch and PennyLane frameworks. 

In addition to the MedMNIST dataset, which is divided into an 8:1:1 ratio, other datasets are typically split into training and testing sets with an 8:2 ratio. 
During the model training phase, we use an Adadelta optimizer with the learning rate of 0.05. 
We use cross-entropy loss and set the batch size as 16.

\subsection{Experimental Results}

\subsubsection{Performance evaluation on the HAM10000}

\begin{figure}[h]
    \centering
    \includegraphics[width=0.60\linewidth]{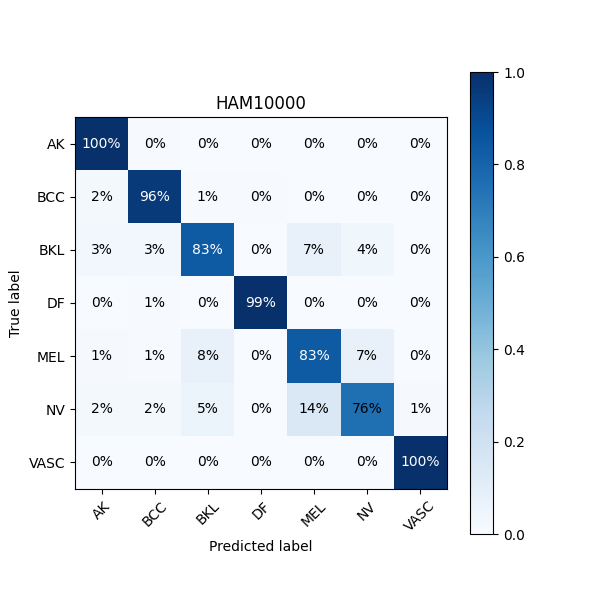}
    \caption{The confusion matrix of the HAM10000}
    \label{fig:matrix of the HAM10000}
\end{figure}

The model's 7-class classification performance on the HAM10000 dataset is presented in \autoref{tab:classifications7}. 
Among all categories, the best performance was demonstrated in DF (dermatofibroma), with a specificity of 99.92\% and both precision and f1-score of 99.47\%. 
For VASC (vascular lesions), the highest recall of 100\% was attained, with specificity, precision, and f1-score all exceeding 98\%. However, the recall for NV (nevus) was relatively low, at only 75.59\%. In \autoref{fig:matrix of the HAM10000}, the confusion matrix for the HAM10000 dataset is shown. It is revealed through analysis that some NV samples were misclassified as MEL (melanoma).
Overall, we achieved an accuracy of 91.14\% on this dataset, demonstrating that the model can perform exceptionally well in multi-class problems, such as the 7-class classification task.

\begin{table}[h]
\centering
\caption{Performance on HAM10000}
\label{tab:classifications7}
\begin{tabular}{ccccc}
\toprule
Categories & Specificity & Precision & Recall & F1-score \\ 
\midrule
AK & 98.65\% & 93.10\% & 99.54\% & 96.21\% \\ 
BCC & 98.95\% & 94.00\% & 96.08\% & 95.03\% \\ 
BKL & 97.74\% & 86.40\% & 83.45\% & 84.90\% \\ 
DF & \textbf{99.92\%} & \textbf{99.47\%} & 99.47\% & \textbf{99.47\%} \\ 
MEL & 96.49\% & 80.05\% & 83.42\% & 81.70\% \\ 
NV & 98.10\% & 86.23\% & 75.59\% & 80.56\% \\ 
VASC & 99.79\% & 98.71\% & \textbf{100\%} & 99.35\% \\ 
\bottomrule
\end{tabular}
\end{table}

\subsubsection{Performance evaluation on the ISIC2017}

\begin{figure}[h]
    \centering
    \includegraphics[width=0.5\linewidth]{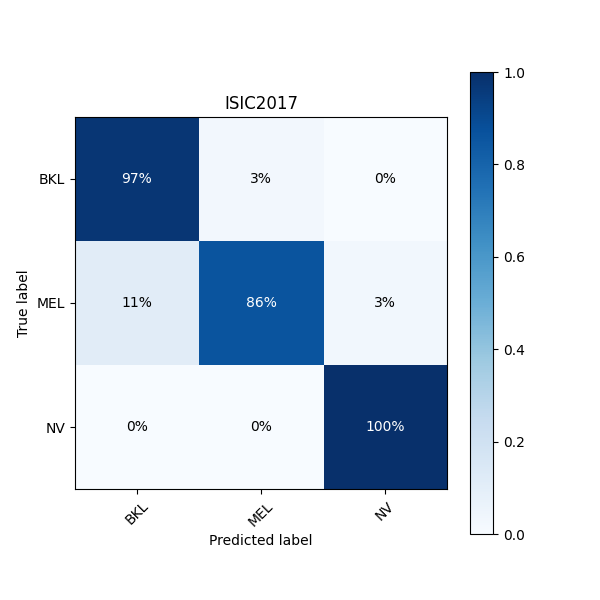}
    \caption{The confusion matrix of the ISIC2017}
    \label{fig:matrix of the ISIC2017}
\end{figure}

The model's 3-class classification performance on the ISIC2017 dataset is presented in \autoref{tab: 3 classifications}. Among the categories, the best performance was observed for NV, with a recall of 100\%, precision of 97.31\%, and f1-score of 98.63\%, the highest across all categories. 
In comparison, MEL (melanoma) achieved a specificity of 98.73\%, but its recall was relatively lower at 85.98\%. 
Furthermore, the precision for BKL (seborrheic) was below 90\%. As shown in the confusion matrix in \autoref{fig:matrix of the ISIC2017}, the prediction accuracy for MEL was 86\%, with 11\% of the samples misclassified as BKL. 
An overall classification accuracy of 94.54\% was achieved. 
Our model demonstrates excellent classification performance, in the context of a three-class classification task.

\begin{table}[h]
\centering
\caption{Performances on ISIC2017}
\label{tab: 3 classifications}
\begin{tabular}{ccccc}
\hline
Categories & Specificity & Precision & Recall & F1-score \\ \hline
BKL & 94.64\% & 89.55\% & 97.35\% & 93.28\% \\ 
MEL & \textbf{98.73\%} & 97.08\% & 85.98\% & C91.19\% \\ 
NV & 98.50\% & \textbf{97.31\%} & \textbf{100\%} & \textbf{98.63\%} \\ \hline
\end{tabular}\end{table}

\subsubsection{Performance evaluation on the MedMNIST}

The results of the model's binary classification task on the MedMNIST dataset are shown in \autoref{tab: 2 classifications}. In the category NL (normal), the specificity and precision are highest, at 97. 94\% and 96. 69\%, respectively. 
However, recall and f1-score are surpassed by the category PNA (Pneumonia). This discrepancy is due to the imbalance in the dataset, where the number of pneumonia cases significantly exceeds that of normal cases, leading to a decline in certain metrics. 

In benchmark experiments, as shown in \autoref{tab: 2 baseline} our model achieved the highest ACC and the second-highest AUC. This demonstrates that our model exhibits a very high correct diagnosis rate for binary classification tasks involving pneumonia.

\begin{table}[h]
\centering
\caption{Performances on MedMNIST}
\label{tab: 2 classifications}
\begin{tabular}{ccccc}
\hline
Categories &Specificity& Precision & Recall & F1-score \\ 
\hline
NL &\textbf{97.94\%}& \textbf{96.69\%}  & 74.79\% & 84.34\% \\ 
PNA &76.77\%& 86.68\% & \textbf{98.46\%} & \textbf{92.20\% } \\ 
\hline
\end{tabular}\end{table}

\begin{table}[h]
\centering
\caption{Benchmark on MedMNIST}
\label{tab: 2 baseline}
\begin{tabular}{ccccc}
\hline
Model & ACC & AUC & \\ 
\hline
ResNet-18 & 86.43\%  & \textbf{95.55\%} \\ 
auto-sklearn & 85.53\%  & 94.18\% \\ classical
AutoKeras & 87.76\%  & 94.72\% \\ 
\textbf{our work} & \textbf{89.58\%}  & 94.87\% \\ 
\hline
\end{tabular}\end{table}

\subsubsection{Comparison experiments and analysis}

Our model was compared with existing models, and the results are presented. These models include CNNs such as ResNet34 (\cite{he2016deep}), DenseNet121 (\cite{huang2017densely}), VGG16+SA (\cite{gao2021soft}), and QCNNs such as Qst-Cla (\cite{ahmed2021classification}), HQCNN (\cite{yadav2023hybrid}), and HQNN-Parallel (\cite{senokosov2024quantum}).
The same dataset split method and a similar training strategy were used to perform comparisons between these models on two medical image datasets related to skin cancer, HAM10000 and ISIC2017.
The results demonstrate that excellent performance was achieved by our designed model on both datasets.

\begin{table}
\centering
\caption{Performance comparison on HAM10000}
\label{tab:Performance comparison on ham10000}
\renewcommand{\arraystretch}{1.3}
\resizebox{\textwidth}{!}{ 
\begin{tabular}{cccccccc}
\hline
Dataset &Model & Accuracy(\%)& Specificity(\%) & Precision(\%) & Recall(\%) & F1-score(\%) & Para (M) \\ \hline
\multirow{7}{*}{HAM10000} 
& ResNet34(\cite{he2016deep}) & 89.35 & 99.71 & 89.18 & 89.33 & 89.14 & 20.32 \\ 
& DenseNet121(\cite{huang2017densely})& 89.50 & \textbf{99.92} & 89.44 & 89.50 & 89.33 & 6.72 \\ 
& VGG16+SA(\cite{gao2021soft}) & 85.53 & 99.46 & 86.00 & 85.42 & 85.49 & 45.75 \\ 
& Qst-Cla(\cite{ahmed2021classification}) & 87.92 & 99.71 & 88.79 & 87.67 & 87.23 & 78.23 \\ 
& HQCNN(\cite{yadav2023hybrid}) & 73.32 & 98.92 & 75.07 & 73.75 & 72.98 & 84.01 \\ 
& HQNN-Parallel(\cite{senokosov2024quantum}) & 82.39 & 98.96 & 83.45 & 82.32 & 82.02 & 11.50 \\
& \textbf{Our work} & \textbf{91.14} & 99.63 & \textbf{91.14} & \textbf{91.09} & \textbf{91.03} & \textbf{2.16} \\ \hline
\end{tabular}
}
\end{table}

On the HAM10000 dataset (\autoref{tab:Performance comparison on ham10000}), an accuracy of 91.14\% was demonstrated by our model, surpassing other models. 
In terms of specificity, nearly 100\% was achieved, indicating exceptional performance in identifying negative class samples. 
Both precision and recall exceeded 91\%, further highlighting the model's strong stability in recognizing positive class samples. 
In comparison, other hybrid quantum models such as Qst-Cla and HQCNN, did not achieve performance metrics above 90\% except for specificity.

\begin{table}
\centering
\caption{Performance comparison on ISIC2017}
\label{tab:Performance comparison on Isic2017}
\renewcommand{\arraystretch}{1.3}
\resizebox{\textwidth}{!}{ 
\begin{tabular}{cccccccc}
\hline
Dataset& Model & Accuracy(\%)& Specificity(\%) & Precision(\%) & Recall(\%) & F1-score(\%) & Para (M) \\ \hline
\multirow{7}{*}{ISIC2017}
& ResNet34(\cite{he2016deep})  & 90.53 & 96.64 & 90.38 & 90.37 & 90.29 & 20.32 \\ 
& DenseNet121(\cite{huang2017densely})& 88.47 & 93.83 & 88.76 & 88.29 & 87.98 & 6.71 \\ 
& VGG16+SA(\cite{gao2021soft}) & 88.59 & 96.26 & 88.45 & 88.49 & 88.39 & 45.73 \\ 
& Qst-Cla(\cite{ahmed2021classification}) & 80.46 & 89.91 & 81.47 & 80.12 & 78.78 & 78.23 \\ 
& HQCNN (\cite{yadav2023hybrid})& 81.31 & 96.82 & 81.22 & 81.06 & 81.13 & 84.01 \\ 
& HQNN-Parallel(\cite{senokosov2024quantum}) & 79.24 & 88.59 & 79.47 & 78.90 & 78.02 & 11.50 \\ 
& \textbf{Our work} & \textbf{94.54} & \textbf{98.50} & \textbf{94.65} & \textbf{94.44} & \textbf{94.37} & \textbf{2.16} \\ \hline
\end{tabular}
} 
\end{table}

On the ISIC2017 dataset (\autoref{tab:Performance comparison on Isic2017}), an accuracy of 94.54\% was reached by our model, significantly outperforming other CNNs such as ResNet34 (90.53\%), DenseNet121 (88.47\%), and VGG16+SA (88.59\%). 
This high accuracy reflects the outstanding overall classification performance of the model, as well as its advantage in extracting complex image features.
Additionally, a specificity of 98.50\% was achieved, showcasing exceptional ability in identifying negative class samples. 
In the evaluation of precision and recall, 94.65\% and 94.44\% were attained, respectively, with an f1-score of 94.37\%. 
These metrics indicate strong performance in identifying positive class samples. 
In comparison, other quantum models, such as HQCNN and HQNN-Parallel, were outperformed, achieving accuracies of 81.31\% and 79.24\%, respectively. 
This difference may be attributed to the complexity of their network structures.
In contrast, our network structure are simpler and more effective.

Furthermore, in terms of number of parameters, only 2.16M parameters are contained in our model, which is significantly smaller than other models. 
While maintaining outstanding performance, significant model size compression has been achieved, making it especially suitable for resource-limited medical imaging tasks. 
This demonstrates that excellent performance can still be achieved in resource-constrained environments.  

\subsubsection{Ablation experiments and analysis}

To evaluate the impact of the quantum circuit splitting technique on the model, ablation experiments were conducted on both medical datasets, HAM10000 and ISIC2017, as shown in \autoref{tab:Ablation study results}.
After adopting distributed computing based on quantum circuit splitting, our model shows an improvement of approximately 2\%-3\% in various metrics. 
This indicates that the introduction of quantum circuit splitting not only overcomes hardware limitations and reduces the required theoretical quantum bits but also enhances the model's learning capability.
This experiment strongly demonstrates the feasibility of the quantum circuit splitting.

\begin{table}[h]
\centering
\caption{Ablation study results}
\label{tab:Ablation study results}
\resizebox{\textwidth}{!}{ 
\begin{tabular}{ccccccc}
\toprule
Dataset & Cut & Accuracy & Specificity & Precision & Recall & F1-score  \\ 
\midrule
\multirow{2}{*}{HAM10000} 
& $\times$ & 89.57\% & 99.65\% & 89.75\% & 89.59\% & 89.50\% \\ 
& $\sqrt{}$ &\textbf{ 91.14\%} & \textbf{99.73\%} & \textbf{91.14\% }& \textbf{91.09\%} & \textbf{91.03\%} \\ 
\midrule
\multirow{2}{*}{ISIC2017} 
& $\times$ & 91.26\% & 98.13\% & 91.26\% & 91.20\% & 91.11\% \\ 
& $\sqrt{}$ & \textbf{94.54\%} & \textbf{98.50\% }& \textbf{94.65\%} & \textbf{94.44\%} & \textbf{94.37\%} \\ 
\bottomrule
\end{tabular}}
\end{table}

\section{Conclusions and Future Work}
\label{Conclusions and future work}

In this paper, a novel distributed hybrid quantum convolutional neural network is presented for the classification of medical images. A quantum convolutional neural network is designed, leveraging the advantages of quantum computing to explore a broader parameter space.
In the context of limited quantum hardware resources, quantum circuit splitting is introduced into the quantum circuit, where the 8-qubit quantum circuit is split into two separate circuits, one 4-qubit and the other 5-qubit, significantly reducing quantum resource usage.
The performance of medical image classification networks is improved by our hybrid quantum convolutional neural network, and the combination of a lightweight classical network and scalable quantum circuits greatly reduces the parameter count and saves computational resources. 
Compared to other CNNs, our work maintains high performance while expanding into the promising domain of quantum computing. When compared to other QCNNs, higher accuracy is achieved by our proposed method, making it more feasible for practical applications.

In future research, we will explore the optimization potential of this model across additional datasets and tasks. The model will be optimized by closely integrating deep learning with parameterized quantum circuits, using fewer quantum bits to simulate larger-scale quantum circuits. Additionally, its application will be extended to tasks beyond image classification.

\section*{CRediT authorship contribution statement}
\textbf{Yangyang Li}: Validation, Supervision, Methodology, Conceptualization. \textbf{Zhengya Qi}: Writing – review \& editing, Writing – original draft, Formal analysis.
\textbf{Yuelin Li}: Writing – review \& editing, Formal analysis.
\textbf{Haorui Yang}: Writing – review \& editing, Data curation. \textbf{Ronghua Shang}: Investigation. 
\textbf{Licheng Jiao}: Investigation.

\section*{Declaration of competing interest}
The authors declare that they have no known competing financial interests or personal relationships that could have appeared to influence the work reported in this paper.

\section*{Acknowledgments}
This work was supported in part by the National Natural Science Foundation of China under Grant 62476209, in part by the Key Research and Development Program of Shaanxi under Grant 2024CY2-GJHX-18, and in part by the Natural Science Basic Research Program of Shaanxi under Grant No.2022JC-45.

\bibliographystyle{elsarticle-harv} 
 \bibliography{reference}

\end{document}